\documentclass[10pt,twocolumn,letterpaper]{article}

\usepackage{cvpr}
\usepackage{times}
\usepackage{epsfig}
\usepackage{graphicx}
\usepackage{amssymb,amsmath}

\usepackage{url}
\usepackage{multirow}
\usepackage{booktabs}
\usepackage[ruled,linesnumbered]{algorithm2e}
\usepackage{mathtools}
\usepackage{tabularx}
\usepackage{varwidth}

\usepackage{fancyhdr}
\DeclarePairedDelimiter{\ceil}{\lceil}{\rceil}
\DeclarePairedDelimiter{\floor}{\lfloor}{\rfloor}

\newcommand{\mbf}{\mathbf{f}}

\newcommand{\mbx}{\mathbf{x}}

\newcommand{\mbK}{\mathbf{K}}

\newcommand{\mbG}{\mathbf{G}}
\newcommand{\mbL}{\mathbf{L}}

\newcommand{\R}{\mathbb{R}}

\newcommand{\calB}{\mathcal{B}}

\newcommand{\calD}{\mathcal{D}}

\newcommand{\calN}{\mathcal{N}}
\newcommand{\calO}{\mathcal{O}}

\newcommand{\calR}{\mathcal{R}}
\newcommand{\calS}{\mathcal{S}}

\newcommand{\calU}{\mathcal{U}}

\newcommand{\calX}{\mathcal{X}}

\def\R{\mathbb{R}}

\def\calX{\mathcal{X}}

\def\det{\mathop{\rm det}\nolimits}

\def\vol{\mathop{\rm vol}\nolimits}

\def\argmin{\mathop{\rm arg\,min}\limits}

\newtheorem{definition}{Definition}

\newtheorem{proposition}{Proposition}

\newcommand{\ignore}[1]{}

\makeatletter
\DeclareRobustCommand\onedot{\futurelet\@let@token\@onedot}
\def\@onedot{\ifx\@let@token.\else.\null\fi\xspace}
\def\eg{{e.g}\onedot} 
\def\ie{{i.e}\onedot}

\def\etal{\emph{et al}\onedot}
\makeatother

\usepackage[pagebackref=true,breaklinks=true,letterpaper=true,colorlinks,bookmarks=false]{hyperref}

\expandafter\def\expandafter\normalsize\expandafter{%
\normalsize\setlength\abovedisplayskip{5.5pt}}

\expandafter\def\expandafter\normalsize\expandafter{%
\normalsize\setlength\belowdisplayskip{5.5pt}}

\cvprfinalcopy 
\ifcvprfinal\fi

\begin{document}

\title{Local High-order Regularization on Data Manifolds}

\author{Kwang In Kim\\
Lancaster University\\
\and
James Tompkin\\
Harvard SEAS\\
\and
Hanspeter Pfister\\
Harvard SEAS\\
\and
Christian Theobalt\\
MPI for Informatics\\
}

\maketitle
\thispagestyle{fancy}

\begin{abstract}
The common graph Laplacian regularizer is well-established in semi-supervised learning and spectral dimensionality reduction. However, as a first-order regularizer, it can lead to degenerate functions in high-dimensional manifolds. The iterated graph Laplacian enables high-order regularization, but it has a high computational complexity and so cannot be applied to large problems. We introduce a new regularizer which is globally high order and so does not suffer from the degeneracy of the graph Laplacian regularizer, but is also sparse for efficient computation in semi-supervised learning applications. We reduce computational complexity by building a local first-order approximation of the manifold as a surrogate geometry, and construct our high-order regularizer based on local derivative evaluations therein. Experiments on human body shape and pose analysis demonstrate the effectiveness and efficiency of our method.
\end{abstract}

\section{Introduction}
The graph Laplacian regularizer is established as one of the most popular regularizers for semi-supervised learning~\cite{ChaSchZie06}, spectral clustering~\cite{ShiMal00,Lux07}, and dimensionality reduction~\cite{BelNiy03}. The underlying assumption for using the graph Laplacian regularizer is that data lie on a low-dimensional sub-manifold, and the object (\eg, a function) of interest should be regularized as defined on the manifold rather than as defined on the entire ambient space. By measuring local pairwise deviations of the function values in the ambient space, the graph Laplacian regularizer approximates the first-order variations on the manifold, thereby enabling us to regularize the function based on its first-order energy without having to know the manifold analytically. 

Despite its solid theoretical background~\cite{BelNiy05,HeiAudLix05} and success in many applications, the graph Laplacian regularizer has an important shortcoming that makes its usage less favorable on data lying in high-dimensional manifolds: as we will discuss, as a first-order regularizer, the null space of the graph Laplacian regularizer contains discontinuous functions on manifolds with dimensionality larger than 2~\cite{NadSreZho09,ZhoBel11}. 

Recently, Zhou and Belkin~\cite{ZhoBel11} proposed an iterated graph Laplacian approach that avoids this \emph{degeneracy} and enables regularization on high-dimensional manifolds. The price for the non-degeneracy and the resulting simplicity of the algorithm is high computational complexity: the iterated graph Laplacian regularizer is constructed by taking powers of the graph Laplacian matrix, which makes the original matrix denser and, accordingly, for large-scale problems (\eg, $O(100,000)$) it cannot be directly applied efficiently.

We propose an empirical regularizer which avoids degeneracy and leads to a sparse matrix. Our algorithm is based on the local linear approximation of the manifold: At each point, the corresponding neighborhood is projected onto its tangent space, where the high-order derivatives of the function are defined in this surrogate geometry. Instead of explicitly calculating high-order derivatives and measuring the corresponding complexity of the function, we measure its reproducing kernel Hilbert space (RKHS) norm. Similar to the graph Laplacian, its sparsity is explicitly controlled based on the local neighborhood structure. We present experimental results on human body shape and pose datasets, which show that our method is superior to graph Laplacian and iterated graph Laplacian techniques in terms of accuracy and computational complexity.

As this paper is equation and symbol rich, we summarize all symbols and notation conventions on the first page of the supplemental material.

\section{Problem statement}
\label{s:problem}
While our proposed regularizer can be used in clustering and dimensionality reduction, as with the graph Laplacian and iterated graph Laplacian regularizers, we focus on \emph{semi-supervised learning} which enables us to compare numerically the performance of each algorithm.

For a set of data points $\calX = \{X_1,\ldots,X_u\}\subset \R^n$ plus the corresponding labels $\{Y_1,\ldots,Y_l\}\subset \R$ for the first $l$ points in $\calX$ where $l\ll u$, the goal of semi-supervised learning is to infer the labels of the remaining $u-l$ data points in $\calX$. Our approach is based on regularized empirical risk minimization:
\begin{align}
 \argmin_{f:\R^n\to\R}\sum_{i=1}^l (Y_i-f(X_i))^2 + \lambda \, \calR(f),
\label{e:learningproblem}
\end{align}
where $\calR(\cdot)$ is the regularization functional. Here, we use the standard squared loss function for simplicity, though our framework is applicable to any convex loss function. This problem can be solved either by reconstructing the underlying function $f$ or by identifying its evaluation $f|_\calX$ on $\calX$. In this paper, we focus on the second case, which is often called \emph{transductive learning}.

Most semi-supervised learning algorithms can be characterized by how the unlabeled data points of $\calX$ are used to construct a corresponding regularizer $\calR(f|_\calX)$. One of the best established regularizers is the graph Laplacian $L$~\cite{Lux07}:
\begin{align}
\calR_L(\mbf):=\mbf^\top L \mbf = \sum_{i,j=1}^u [W]_{ij}(f_i-f_j)^2,
\label{e:lapenergy}
\end{align}
where $f_i=f(X_i)$, $\mbf:=f|_\calX=[f_1,\ldots,f_u]^\top$, and $W$ is a non-negative input similarity matrix which is typically defined based on a Gaussian:
\begin{align}
[W]_{ij}=\exp\left(-\frac{\|X_i-X_j\|^2}{b}\right).
\label{e:lapgaussian}
\end{align}

One way of justifying the use of the graph Laplacian comes from its limit case behavior as $u\rightarrow\infty$ and $b\to 0$: When the data $\calX$ is generated from an underlying manifold $M$ with dimension $m\leq n$, \ie, the corresponding probability distribution $P$ has support in $M$, the graph Laplacian converges to the Laplace-Beltrami operator $\Delta$ on $M$ \cite{BelNiy05,HeiAudLix05}. The Laplace-Beltrami operator can be used to measure the first-order variations of a continuously differentiable function $f$ on $M$:
\begin{equation}
\|f\|_\Delta^2:= \int_M f(X) [\Delta f|_X] dV(X)= \int_M \|\nabla f|_X\|_g^2 dV(X),
\label{e:lapnorm}
\end{equation}
where $g$ is the \emph{Riemannian metric}, and $dV$ is the corresponding \emph{natural volume element}~\cite{Lee97} of $M$. The second equality is the result of Stokes' theorem. Accordingly, a graph Laplacian-based regularizer $\calR_L$ can be regarded as an empirical estimate of the first-order variation of $f$ on $M$ based on $\calX$.

\begin{figure*}[t]
\includegraphics[width=\textwidth]{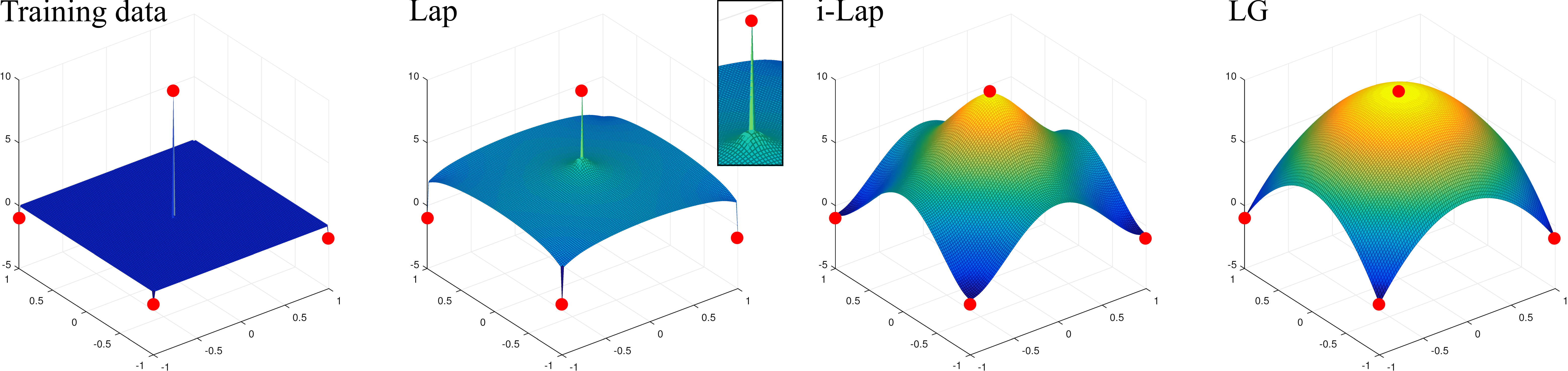}
\caption{\label{f:toy2d}Example on 2D data. Section 5 contains details of this toy example; the surface in the training data plot is to help with visualization only, and no regularization has taken place. The \emph{Lap} result largely fails to regularize, apart from points very near to the original training data. These spikes can be seen in the zoom inlay. The result of \emph{i-Lap} looks \emph{hyperbolic} because its null space includes polynomials. In this example, both \emph{i-Lap} and \emph{LG} are acceptable since they lead to smooth functions. Inspired by \protect\cite{ZhoBel11}.}
\end{figure*}

However, the convergence of the graph Laplacian $L$ to the Laplace-Beltrami operator $\Delta$ reveals an important shortcoming for it to be used as the standard regularizer for high-dimensional data: For high-dimensional manifolds ($m>1$), the null space of $\Delta$ includes discontinuous functions on $M$. This is suggested by the \emph{Sobolev embedding theorem} that states that, in general, any (semi-)norm induced by differential operators with order $d\leq m/2$ will have discontinuous functions in its null space~\cite{Ros08}. In particular, the norm $\|\cdot\|_\Delta$ in Eq.~\ref{e:lapnorm} which measures the first-order variation has a null space consisting only of continuous functions (in particular, constant functions) when $m=1$ only. For $m>1$, the null space of $\Delta$ contains some discontinuous functions as a subset of $L^2$ space which are equivalent almost everywhere to constant functions, except for the set of \emph{measure zero}~\cite{dud02}. In other words, there are ``spiky'' functions $f$, \eg, Dirac delta functions, with norm $\|f\|_\Delta^2=0$ (Fig.~\ref{f:toy2d}). 

This is especially important in semi-supervised learning because we actively minimize the regularized risk of attaining a zero value by such a function (Eq.~\ref{e:learningproblem}). While this has been well-known in statistics, its effect on semi-supervised learning has only recently been analyzed by Nadler~\etal~\cite{NadSreZho09}. They showed that, in the limit case (\ie, $u\to \infty$), where $\calR_L$ is used, indeed the null space of the empirical risk functional (Eq.~\ref{e:learningproblem}) includes a function $f$ which is zero everywhere except for the labeled data points $\{X_1,\ldots,X_l\}$, where $f$ agrees with the given labels, and no generalization is obtained.

In practice, due to the finite number of data points $u$, the learned function $f$ (more precisely, its evaluation $\mbf$ on $\calX$) is not a Dirac delta function exactly, but is a very steep, sheer-sided spike which peaks at the labeled data points (Fig.~\ref{f:toy2d}). For discrete problems, \eg, classification, where only relative values of $f$ are relevant, it is possible to normalize the output values based on the local distribution of $f$ to soften such peaks, as exemplified in \cite{WuLiCha13}. However, this technique is not applicable for learning continuous functions.

Zhou and Belkin~\cite{ZhoBel11} presented the first approach that explicitly prevents this degenerate case in semi-supervised learning. They proposed using powers of graph Laplacian (or \emph{iterated graph Laplacian}) as a regularizer:
\begin{align}
\calR_{L^p}(\mbf):=\mbf^\top L^p \mbf,
\label{e:iterlap}
\end{align}
with $p>\frac{m}{2}$. In the limit case as $u \rightarrow \infty$, $L^p$ converges to $\Delta^p$, which corresponds to the penalizer of (selected) \mbox{$\ceil{\frac{p}{2}}$-th} order variations in the context similar to Eq.~\ref{e:lapnorm}~\cite{ZhoBel11}:
\begin{equation}
\|f\|_{\Delta^p}^2= \int_M f(X) [\Delta^p f|_X] dV(X),
\label{e:lappnorm}
\end{equation}
which is infinite when $f$ is discontinuous. The ability to regularize over higher-order derivatives avoids the degenerate case of learning discontinuous functions.

One of the major limitations of iterated graph Laplacian is that, due to the density of the resulting matrix $L^p$, it cannot be directly applied to large-scale problems. For a non-iterated graph Laplacian, finding the minimizer of Eq.~\ref{e:learningproblem} with $\calR_{L}$ requires building and solving a linear system of size \mbox{$u\times u$}. Even for large-scale problems (\eg, $u\approx 10^5$), this is affordable since the corresponding weight matrix $W$ can be well-approximated by a sparse matrix constructed from a $k$-nearest neighbor (NN) graph. However, in general, iterating $L$ (taking powers $L^p$) makes a sparse matrix denser. This is especially true when $p$ is large, which is required for high-dimensional data, as suggested by the Sobolev embedding theorem. For instance, with $u=50,000$, solving Eq.~\ref{e:learningproblem} with iterated graph Laplacian is $15\times$ slower (Sec.~\ref{sec:experiments}) than the Laplacian case.

\section{Local high-order regularization}
\label{s:lapalt}
Our goal is to build a new regularizer that shares the desirable properties of both penalizing discontinuous functions with $L^p$ and being sparse in $L$ for fast computation. To achieve this goal, we build a global regularization matrix $G$ based on local regularizers evaluated at each point in $\calX$.

First, we take a class of high-order manifold operators as regularizers by adopting the regularization framework of Yuille and Grzywacz~\cite{YuiGrz88}. These regularizers correspond to generalizations of Eq.~\ref{e:lapnorm}:\footnote{As a special case, when $c_{p}=1$ and $\{c_k\}_{k\neq p}=0$, $\|\cdot\|_D^2$ becomes $\|\cdot\|_{\Delta^{p}}^2$ (Eq.~\ref{e:lappnorm}). In general, different choices of differential operators are possible, \eg, Hessian, rather than the powers of $\Delta$ and $\nabla$. This choice was motivated by the demonstrated empirical success of the resulting regularizer in many applications~\cite{YuiGrz88}, and the computational efficiency as facilitated by the use of the corresponding Gaussian RKHS as discussed in Sec.~\ref{s:LGR}.}
\begin{align}
\|f\|_D^2:=  \int_M \sum_{k=1}^\infty c_k |D^k f|_X|^2 dV(X),
\label{e:lapnormser}
\end{align}
\begin{equation}
D^{k}f = 
\left\{ \begin{aligned}
\Delta^k f, 		&\hspace{1cm} \text{for even } k \\
\nabla(\Delta^k f), 	&\hspace{1cm} \text{for odd } k 
\end{aligned} \right.
\label{e:dervopr}
\end{equation}
\begin{equation}
|D^{k}f|^2 = 
\left\{ \begin{aligned}
(D^{k} f)^2, 		&\hspace{1cm} \text{for even } k \\
g(D^{k} f,D^{k} f), 	&\hspace{1cm} \text{for odd } k 
\end{aligned} \right.
\end{equation}
where $k$ is the order of the derivative operator, and coefficients $c_k\geq 0$.

For a known manifold with known metric and \emph{Christoffel symbols}~\cite{Lee97}, the derivative operators in Eq.~\ref{e:dervopr} are easy to calculate. However, in most practical applications, the manifold is not directly observed but is only indirectly observed as a point cloud of sampled data points $\calX\subset \R^n$, where $M$ is a ($m$-dimensional) sub-manifold of $\R^n$. Accordingly, direct calculation of Eq.~\ref{e:dervopr} is infeasible.

\paragraph{A local first-order approximation $D_0$.} We bypass this problem by using a local first-order approximation $T_X(M)$ of manifold $M$ at each point $X$ ($M_X$) in $\R^n$ as a proxy geometry for $M$ near $X$. 
Since $T_X(M)$ is identified with $\R^m$, evaluating the derivative operators in Eq.~\ref{e:dervopr} on $X$ boils down to the calculation of the derivative operators in Euclidean geometry. In particular, evaluating the Laplace-Beltrami operator becomes the calculation of the Laplacian operator:
\begin{align}
D^2_0 f|_X=\Delta_0 f|_X=\sum_{r=1}^m\partial_r^2  f|_X.
\label{e:laprc}
\end{align}
Subscript $0$ denotes operators defined on the proxy geometry, where $\Delta_0[\cdot]|_X$ is the Laplacian defined at $T_X(M)$. $\partial_r$ is shorthand for $\frac{\partial}{\partial x^r}$. Practically, the dimension of $m$ is unknown and so is a hyper-parameter.

With a manifold approximation, the next step is to construct approximations of Eq.~\ref{e:dervopr} and Eq.~\ref{e:laprc} given $\calX$ and $f|_\calX$. Suppose that for each data point $X_i$, the corresponding $k$-NN $N_k(X_i)\subset \calX$ are identified. First, we estimate the first-order approximation $T_{X_i}(M)$ by performing principal component analysis on $N_k(X_i)$~\cite{DonGri2003}: The representations $\{\mbx_j\}_{j=1}^k$ of $N_k(X_i)$ on $T_{X_i}(M)$ are given as the first $m$-principal components of $N_k(X_i)$. 
Then, at $X_i$, the approximation of the Laplacian in Eq.~\ref{e:laprc} is obtained by fitting a smooth interpolation $\varphi^i$ in $(x)$ to $\{f(X_j)\}_{j=1}^k$ and then extracting the trace of the resulting Hessian $H \varphi^i$ of $\varphi^i$, which we denote as $S^{(2)}(X_i)$. The surrogate function $\varphi^i$ can be a (constrained) second-order polynomial $h^i$ (for $\Delta$) or a Gaussian kernel interpolation $q^i$ (for $\Delta^k$, $k>0$):
\begin{align}
h^i(\mbx) &= f(X_i) + \sum_{r=1}^m [a^i]_{r} x^r+ \sum_{r=1,s=r}^{m} [b^i]_{r,s} x^rx^s,\\
q^i(\mbx) &= f(X_i) + \sum_{j=1}^k [\alpha^i]_jK(\mbx_j,\mbx),
\label{eq:gpregr}
\end{align}
where $\mbx = [x^1,\ldots,x^m]^\top$, and 
\begin{align}
\label{e:gaussian}
K(\mbx,\mbx')=\exp\left(-\frac{\|\mbx-\mbx'\|^2}{\sigma^2}\right).
\end{align}
The coefficients $\{a^i,b^i\}$ and $\{\alpha^i\}$ of $h^i$ and $q^i$, respectively, are calculated as the standard least squares fit:
\begin{align}
[a^i,b^i]&=\argmin_{w \in \R^{m+m(m+1)/2}} \sum_{j=1}^k \Big(f(X_j) - h^i(\mbx_j)\Big)^2,\\
\alpha^i&=\hspace{0.5cm}\argmin_{a \in \R^k} \hspace{0.55cm}\sum_{j=1}^k \Big(f(X_j) - q^i(\mbx_j)\Big)^2,
\label{e:gpenergy}
\end{align}
where $w$ is a vector of linear and quadratic coefficients in the second-order polynomials.

By combining these estimates of the local Laplacians and re-arranging the variables, one can construct a matrix $B$ as a new regularizer on a point cloud $\calX$:
\begin{align}
\| f\|_{\Delta_0}^2\approx\calR_{B}(\mbf)=\mbf^\top B\mbf = \sum_{i=1}^u f(X_i) S^{(2)}(X_i).
\label{e:globalmat}
\end{align}

To evaluate the squared Laplacian operator $\Delta_0^2|_{X_i}$, we calculate the corresponding fourth-order derivatives of $\varphi$. In the case when $\varphi=q$, the derivatives of $\varphi$ of any order are easily calculated by noting that the derivative of a Gaussian function can be evaluated based on the original Gaussian and the combinations of Hermite polynomials~\cite{Kar09}. The corresponding empirical regularizer $\calR_{E}$ based on a finite number of points $\calX$ can be constructed similarly to Eq.~\ref{e:globalmat}:
\begin{align}
\calR_{E}(\mbf)=\sum_{k=1}^\infty c_k \mbf^\top E^{(k)}\mbf:=\sum_{i=1}^u \calS_{X_i}(f),
\label{e:regulonx}
\end{align}
where $k$ indexes the order of derivatives, $\calS_{X_i}(f)=\sum_{k=1}^\infty  c_k |S^{(k)}(X_i)|^2$, $S^{(k)}(X_i)$ corresponds to an empirical approximation of $D^kf|_{X_i}$, and $E^{(k)}(X_i)$ is the corresponding regularization matrix.

\paragraph{Summary} Our regularizer $\calR_E$ is constructed by combining a set of local high-order regularizers, each of which is obtained based on a local first-order approximation of $M$. This avoids explicit calculation of high-order derivatives on $M$. Our regularizer $\calR_E(\mbf)$ is explicitly given as a sparse matrix $E$, \ie, $\calR_E(\mbf)=\mbf^\top E \mbf$, where $E$ is obtained by aligning the local matrices $\{E^{(k)}\}$. Since this is a combination of local high-order regularizers, it is a global high-order regularizer, and therefore it avoids the degeneracy of the graph Laplacian regularizer. As a combination of local high-order regularizers, $\calR_E$ is a global high-order regularizer, and therefore it avoids the degeneracy of the graph Laplacian regularizer.

Explicitly calculating $\{E^{(k)}\}$ is both numerically unstable and computationally demanding. Therefore, we propose a stable approximation of $\calR_E$ in Sec.~\ref{s:LGR}. Before we explain this more-practical implementation, for interested readers, we discuss the relationship between the operators $D$ and $D_0$.

\subsection{Relation between $D$ and $D_0$.}
The regularizer $\calR_{E}$ depends on the local first-order approximation $T_X(M)$ at each $X$. If the $M$ is smoothly embedded in the ambient space $\R^n$, especially in the sense that the corresponding \emph{second fundamental form}~\cite{Lee97} is bounded, then the approximation error is third-order: Let $d_X:=d_X(\cdot,\cdot)$ be the geodesic distance between two points on $M$ in the neighborhood $\calN(X)$ of $X$,\footnote{The \emph{injectivity radius} $\text{inj}(X)$ of $X\in M$ is always positive~\cite{Lee97}. Here, we assume that $\calN(X)\subset\text{inj}(X)$.} then the distance $\tilde{d}_X$ between these points in the proxy geometry $T_X(M)$ is related as~\cite{BelNiy05,HeiAudLix05}
\begin{align}
d_X=\tilde{d}_X+\calO(d_X^3).
\end{align}

The use of local first-order approximations to a manifold is justified by its success in many applications (\eg, \cite{RowSau00,DonGri2003}). We support this approximation further by noting that the corresponding orthonormal coordinates in $T_X(M)$ can be regarded as approximations of \emph{Riemannian normal coordinates}~\cite{KimSteHei10}. In a Riemannian normal coordinate chart centered at a point $X$, the manifold appears Euclidean up to second-order. Specifically, at $X$, the corresponding Riemannian metric $g$ becomes Euclidean: the first order derivatives vanish, and evaluating the Laplace-Beltrami operator boils down to the calculation of the Laplacian in Euclidean space:
\begin{align}
\Delta f|_X=\sum_{r,s=1}^m\frac{\partial_r(g^{rs}\sqrt{\det{g}}\partial_s f)}{\sqrt{\det{g}}}=\Delta_0 f|_X,
\end{align}
where $\partial_r=\frac{\partial}{\partial x^r}$, $\delta^r_s=\sum_{t}g^{rt}g_{ts}$, $\delta^r_s$: $\delta^r_s=1$ if $r=s$ and $0$, otherwise, $g_{rs}=g(\partial_r,\partial_s)$, and $\det{g}$ is the determinant of the matrix evaluation $\{g_{rs}\}$. Using this setup, similarly to the graph Laplacian $L$ case, one can show the convergence of the matrix $B$ (Eq.~\ref{e:globalmat}) to $\Delta$ in the limit case as $u\to\infty$, the diameter $\epsilon$ of $N_k$ is controlled carefully: 

\begin{definition}[Audibert and Tsybakov~\cite{AudTsy07}]
For given constants $c_0,\epsilon_0>0$, a Lebesgue measurable set $A\subset \mathbb{R}^m$ is called $(c_0,\epsilon_0)$-\emph{regular} if
\begin{align}
\lambda[A \cap \mathcal{B}(\mathbf{x},\epsilon)]\geq c_0 \lambda[\mathcal{B}(\mathbf{x},\epsilon)], \ \forall \epsilon\in [0,\epsilon_0],\forall \mbx\in A,\nonumber
\end{align}
where $\lambda[S]$ is the Lebesgue measure of $S\subset \mathbb{R}^m$~\cite{dud02}. We fix constants $c_0,\epsilon_0>0$ and $0<\mu_{\min}<\mu_{\max}<\infty$ and a compact $\mathcal{C}\subset \mathbb{R}^d$. We say that the \emph{strong density assumption} is satisfied if the distribution $P$ is supported on a compact $(c_0,\epsilon_0)$-regular set $A \subseteq\mathcal{C}$ and has a density $\mu$ w.r.t.~$\lambda$ bounded above and below by between $\mu_{\text{min}}$ and $\mu_{\text{max}}$
\begin{align}
\mu_{\min}\leq\mu(\mbx)\leq\mu_{\max}, \  \forall \mbx\in A \text{ and }\mu(\mbx)=0 \text{ otherwise}.\nonumber
\end{align}
\end{definition}

\begin{proposition}
If Hessian $Hf$ on $M$ is Lipschitz continuous with the Lipschitz constant $\gamma$, and the natural volume element $dV$ is bounded in the sense that the underlying probability distribution $P$ satisfies strong density assumption, then there are constants $C_1,C_2,\mu_0>0$ such that with probability larger than $1-(m^2+3m)\exp(-C_2u\epsilon^m)$:
\begin{equation}
\label{e:dev}
|tr[Hh(\mbx)]-\Delta f(X)|^2 \leq \frac{k}{u\epsilon^m} \frac{C_1\epsilon^2\gamma^2}{\mu_0},
\end{equation}
where $tr[A]$ calculates the trace of $A$,  $k=|\calX \cap \calB(X,\epsilon)|$, and $\calB(X,\epsilon)$ is the $\epsilon$-neighborhood of $X$ in coordinates, \ie $\calB(X,\epsilon):=\{X':\|\mbx-\mbx'\|_{T_X(M)}\leq \epsilon\}$, with $\mbx'$ being the coordinate representation of $X'$.
\end{proposition}
The proof of this convergence be found in the supplemental material. For simplicity of proof, we use the $\epsilon$-neighborhood $\calB(X,\epsilon)$ instead of $k$-NNs $N_k(X)$. It can be easily modified for the $k$-NN case (see supplemental material). Accordingly, in Eq.~\ref{e:dev}, $\epsilon$ is the only parameter to be controlled to obtain the convergence. The role of $\epsilon$ is similar to the width of the Laplacian \emph{weight function} (Eq.\ref{e:lapgaussian}) in \cite{BelNiy05}: Roughly, decreasing $\epsilon$ guarantees that the local surrogate function $h$ is flexible enough to well-approximate $f$. However, it should not shrink too fast to ensure that there are sufficient data points $k$ in $\calB(X,\epsilon)$ to prevent $h$ from \emph{overfitting} to $f$. This leads to the condition that \hbox{$\epsilon^m$-shrink} should be slower than $u$-increase, so that $u\epsilon^m\to \infty$. The number of neighborhoods $k$ in Eq.~\ref{e:dev}, given as $|\calB(X,\epsilon)\cap\calX|$, is automatically controlled by sampling $\calX$ from $P$. This leads to $\calO(\frac{k}{u\epsilon^m})=1$ (see supplemental material) guaranteeing quadratic ($\epsilon^2$) convergence. All other constants $C_1$, $C_2$, $\mu_0$, and $\gamma$ are independent of $u$. 

The strong density assumption is moderate. In particular, it holds for any compact manifold with a continuous distribution.

In general, the derivatives of the metric $g$ with orders higher than 2 are non-vanishing even in normal coordinates. In this case, for instance, $\Delta_0^2 f|_X$ deviates from $\Delta^2 f|_X$ in third-order:
\begin{align}
\Delta^2 f|_X=\Delta_0^2 f|_X+\calD^3(f|_X),
\end{align}
where $\calD^3(f|_X)$ contains selected derivatives of $f$ at $X$ up to third-order.\footnote{This can be easily verified by expanding the derivatives in normal coordinates at $X$:
\begin{align}
\Delta^2 f=\sum_{i,j,r,s=1}^m \Bigg(&\partial_i\partial_j [g^{rs}\partial_r\partial_s f]+\partial_i\partial_j\left[\partial_r[\partial g^{rs}]\partial_s f\right]\nonumber\\
&+ \frac{1}{2}\partial_i\partial_j\left[g^{rs}\sum_{t,u=1}^m g^{t u}\partial_r[\partial g_{t u}]
  \partial_s f\right]
 \Bigg).\nonumber
\end{align}
}

However, since they agree at the highest (fourth) order, $\Delta_0^2$ shares two important properties with $\Delta^2$ which are precisely what leads to a \emph{proper} regularizer for $m<4$. When $m<4$, and the metric $g$ and the embedding $\hat{i}:M\to \R^n$ are smooth:
\begin{enumerate}
\item $c_2\Delta_0+c_4\Delta_0^2$ with $c_2,c_4>0$, has the null space consisting of truly constant functions (\ie, excluding the degenerate functions which deviate from constant functions on sets of measure zero), and 
\item The evaluation of the corresponding norm defined similarly to Eq.~\ref{e:lapnorm} is infinite for any discontinuous functions. 
\end{enumerate}

This property extends to general high-order cases: The approximation error of $\Delta^k_0|_X$ to $\Delta^k|_X$ is of order $k-1$ and, for a manifold with dimension $m\geq 4$, the regularizers $\|\cdot\|_{D_0}^2$ that replaces $D^k$ with $D^k_0$ in $\|\cdot\|_D^2$ (Eq.~\ref{e:lapnormser}) with $c^1,\ldots,c^{\floor{m/2+1}}>0$ share the same null space with \mbox{$\|\cdot\|_D^2$}. Furthermore, their evaluations on any discontinuous functions produce infinite value.

\section{Local Gaussian regularization}
\label{s:LGR}
The regularization cost functional $\calR_{E}$ (Eq.~\ref{e:regulonx}) has both the desired properties of being a high-order regularizer and of leading to a sparse system. However, evaluating it requires explicitly calculating the powers of the Laplacian evaluation $\Delta_0^kf|_{X_i}$ at each point $X_i\in\calX$ and for each non-zero coefficient $c_k$. This is not only tedious but also numerically unstable since, in practice, the corresponding high-order derivatives are estimated by fitting a function $\varphi^i$ to only a small number ($k$) of data points $N_k(X_i)$: fitting a high-order polynomial (as an extension of $h^i$ in Eq.~\ref{eq:gpregr}) is very unstable in general. While this can be resolved with smooth Gaussian interpolation \ie $\varphi^i=q^i$, due to the existence of high-order polynomials contained in the derivatives of $q^i$ (Eq.~\ref{eq:gpregr}), the resulting derivative estimates can still be unstable, \ie, perturbed significantly with respect to slight variations of $f$.

We focus on a special case of the regularization functional $\calR_{E}$, with a specific choice of derivative operator contribution $\{c_k\}$, which enables us to bypass the explicit evaluation of individual derivatives $D^k$ while retaining the desired properties of being a sparse, robust, and high-order regularizer.

First, the stability problem in evaluating derivatives can be addressed by taking integral averages of derivative evaluations ($D^kf$; Eq.~\ref{e:dervopr}) and the corresponding magnitude $|D^kf|$ within a neighborhood $\calU(X_i)$ of $X_i$, rather than their point evaluations at $X_i$. For instance, for derivative operators of even powers, instead of $|D^{2k}_0 f|_{X_i}|$ (Eq.~\ref{e:lapnormser}), we use:
\begin{align}
|\widetilde{D}^{2k}_0 f|_{X_i}|=\frac{1}{\vol(\calU({X_i}))}\int_{\calU({X_i})}[\Delta^k_0  \varphi^i|_\mbx]^2  d\mbx,
\label{e:inftydiff}
\end{align}
where $\vol(A)$ measures the volume of $A \subset T_{X_i}(M)$, which is a fixed constant given $M$.

This still requires explicit calculation of derivatives. However, for the special case of Eq.~\ref{e:lapnormser} where the coefficients $\{c_k\}$ are given as:
\begin{align}
c_k &= \frac{\sigma^{2k}}{k!2^k},
\label{e:inftycoeff}
\end{align}
with $\sigma^2$ as defined in (\ref{e:gaussian}) we can efficiently calculate an approximation: First, the \emph{local energy} of $\varphi^i=q^i$ over $T_{X_i}$ defined as
\begin{align}
\|q^i\|_D^2:= \sum_{k=1}^\infty c_k \int_{T_{X_i}(M)} |D^k q^i|_\mbx|^2 d\mbx=\|q^i\|_K^2,
\label{e:gnorm}
\end{align}
can be analytically evaluated as the corresponding Gaussian reproducing kernel Hilbert space (RKHS) norm $\|\cdot\|_K$: The second equality is one of the central results in regularization theory~\cite{YuiGrz88}, established by obtaining $q^i$ as the solution of a minimization that combines the energy in Eq.~\ref{e:lapnormser} with an empirical loss in Eq.~\ref{e:gpenergy}. This is always possible as $q^i$ has $k$ degrees of freedom, and leads to an Euler-Lagrange equation that renders $k$ as Green's function of our operator $D$.

Second, we note that, for large $u$, the local energy (Eq.~\ref{e:gnorm}) well approximates the sum of local stabilized derivations (Eq.~\ref{e:inftydiff}). For a Gaussian function $K(\mbx_j,\cdot)$, its value and derivatives decrease rapidly as the corresponding points of evaluation deviate from center $X_j$ (depending on its \emph{width} $\sigma^2$). Accordingly, its support is \emph{effectively} limited within a neighborhood $\calU'(X_j)$. Since $D^k q^i$ is a kernel expansion of $N_k(X_i)$, its support is limited to a larger neighborhood $\calN(X_i)$ of $X_i$ that encompasses $\{\calU'(X_j)$, $\forall X_j\in N_k(X_i)\}$. Then, we set $\calU({X_i})$ by $\calN(X_i)$ and obtain the local energy $\|q^i\|_D^2$ as a replacement of the integrand in (\ref{e:lapnormser}).

In general, for given $\calU({X_i})$, this approximation becomes more accurate as $\sigma^2$ and $N_k(X_i)$ decrease to zero, which is the case as $u\to\infty$ (see accompanying supplemental material). However, for practical applications, we do not tune $\sigma^2$ or $N_k(X_i)$ to minimize error or to achieve a desired level of accuracy since explicitly calculating the corresponding error is tedious (see Appendix). More importantly, having too small $\sigma^2$ or $N_k(X_i)$ for finite $u$ will lead to a bad interpolation function: a Gaussian kernel interpolation with small $\sigma^2$ may lead to a highly non-linear function $q^i$ that overfits to $\{f(X_j)\}_{j=1}^k$. While we propose setting $\sigma^2$ and $N_k(X_i)$ as decreasing functions with respect to $u$ so that the approximation becomes exact as $u\to\infty$, for practical applications with fixed $u$ (including our experiments), we implicitly determine the diameter of $N_k(X_i)$ based on the selected $k$-NN, and regard $k$ and $\sigma^2$ as hyper-parameters. As described in Sec.~\ref{sec:experiments}, $\sigma^2$ is actually adaptively determined based on $N_k(X_i)$ and accordingly only $N_k(X_i)$ is tuned.

Now, we build a new regularizer $\calR_G$ as a combination of local regularizers on $\varphi^i-f(X_i)$ for $i=1,\ldots,u$, similarly to Eq.~\ref{e:regulonx} in Section \ref{s:lapalt}:
\begin{align}
\calR_G(\mbf)=\sum_{i=1,\ldots,u}{\mbf^{i}}^\top \mbG^i \mbf^{i}
\label{e:gorgreg}
\end{align}
with:
\begin{align}
{\mbf^{i}}^\top \mbG^i \mbf^{i}&=\|f(X_i)-\varphi^i(\cdot)\|_K^2\\
&={\mbf^{i}}^\top(I-\mathbf{11}^i)^\top {(\mbK^{i})}^{+}(I-\mathbf{11}^i){\mbf^{i}},
\label{e:localregl}
\end{align}
where $[\mbK]_{lm}=K(\mbx_l,\mbx_m)$, $\mbf^{i}=[f(X_1),\ldots,f(X_k)]^\top$, $\mbK^+$ is the Moore-Penrose pseudoinverse of $\mbK$, and $\mathbf{11}^i$ is an indicator matrix whose element is zero except for the $l(i)$-th column that consists of ones with $l(i)$ being the index of $X_i$ in $N_k(X_i)$.

\section{Augmenting null spaces}
\label{s:augnull}
\begin{figure*}
\centering
\includegraphics[width=\linewidth]{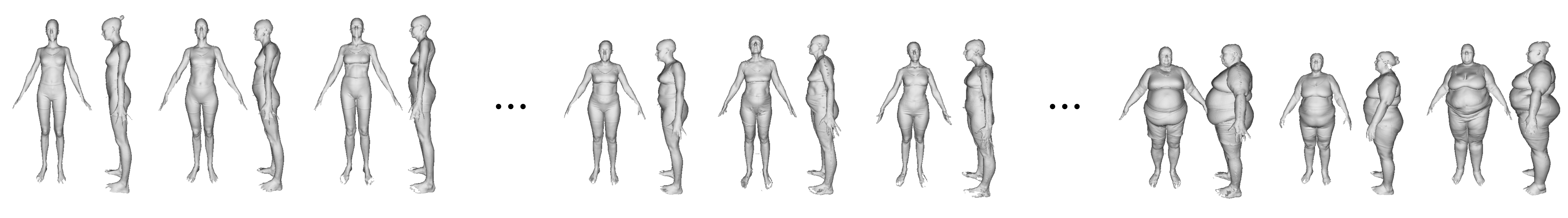}
\caption{The CAESAR database contains $4,258$ 3D scans of human beings, along with ground-truth body measurements taken with calipers. Here, we see variation in female shape across the database.}
\label{fig:caesar}
\end{figure*}

Our local Gaussian regularizer completely eliminates the possibility of generating degenerate functions and so provides a valid regularization on high-dimensional manifolds. Further, it is designed as a combination of \emph{local} regularizers (Eq.~\ref{e:gorgreg}) and so is tailored to incorporate a priori knowledge of the local behavior of functions. In particular, it is easy to tune the regularizer such that it does not penalize functions with desirable properties (\ie, to augment the null space of the regularizer so that it contains those functions). One good choice for $\mbf$ are geodesic functions: both Donoho and Grimes \cite{DonGri2003} and Kim et al.~\cite{KimSteHei10} have demonstrated that \emph{geodesic functions}, which are linear along geodesics, i.e., nothing more than linear functions in Euclidean space, are preferred over other functions since they correspond to the most \emph{natural} parametrization of the underlying data. 

The geodesic functions are completely characterized by their local behavior. In particular, in the Riemannian normal coordinates, they are locally linear functions. Accordingly, we can easily add geodesic functions to the null space of the global regularizer $\calR_G(\mbf)$ by including linear functions in the null space of the local regularizers (Eq.~\ref{e:localregl}): We fit a linear function to $\mbf^i$ and \emph{subtract} the resulting function from $\mbf^i$ before we fit the non-linear function (Eq.~\ref{eq:gpregr}). This can be easily incorporated into new local regularization matrices: 
\begin{align}
 (\mbG')^i  &=\|f(X_i)-\varphi^i_L(\cdot)-\varphi^i(\cdot)\|_K^2\\
&=(\mbL^i)^\top {(\mbK^{i})}^{+}\mbL^i,\label{e:locallinearnull}
\end{align}
where $\varphi^i_L(\cdot)$ is the linear regressor fitting $\mbf^i$ in normal coordinates (\ie, $\varphi^i_L(\mbx)=(\Phi^i_L)^+(I-\mathbf{11}^i)\mbf^i\mbx$), $\Phi^i_L \in \R^{k \times m}$ is the design matrix whose rows correspond to the normal coordinate values of $N_k(X_i)$, and
\begin{align}
\label{e:li}
\mbL^i=I-\mathbf{11}^i-\Phi^i_L(\Phi^i)_L^+(I-\mathbf{11}^i).
\end{align}

The new regularization functional $\calR_{G'}$, in which $\{(\mbG')^i\}$ replaces $\{\mbG^i\}$, has a richer null space: a one-dimensional space of constant functions plus an $m$-dimensional space of geodesic functions. This null space should not be confused with the \emph{too large} null space of the original graph Laplacian regularizer. The null space of our updated local Gaussian regularizer does not include any degenerate functions. 

While this setup does not cause any noticeable increase of computational complexity, in our preliminary MoCap experiments (see Sec.~\ref{sec:experiments}), this reduced error rates by around $3\%$. Accordingly, throughout the entire experiments, we use this new local Gaussian regularizer.

$\calR_{G'}$ construction pseudocode is in Algorithm \ref{a:mainalg}. Supplemental MATLAB code is available on the author's webpage. This real code references the pseudocode to aid explanation.
\begin{algorithm}[t]
\caption{The construction of the regularization functional $\calR_{G'}$ from a point cloud $\calX$.}
\SetKwInput{Input}{Input}
\SetKwInput{Output}{Output}

\Input{$\calX=\{X_1,\ldots,X_u\}$, manifold dimension $n$, $k$.}
\Output{{$G'$.}}
\BlankLine
Initialization: Find $k$ nearest neighbors, \eg, build KD-tree\;
\For{$i=1,\ldots,u$}
{
	Construct the local approximation $M$ at $X_i$ using $n$-dimensional PCA of $N_k(X_i)$\;
	Calculate the local regularization matrix $\mbG^i$ for $N_k(X_i)$ in the PCA representation: $(\mbG')^i=(\mbL^i)^\top (\mbK^i)^{+1}\mbL^i$ (Eqs.~\ref{e:locallinearnull} and \ref{e:li})\;
}
Re-arrange $\{(\mbG')^i\}$\ according to the indices of $\{\mbf^i\}$ in $\mbf$ to construct matrix $G'$ s.t. $\mbf^\top G' \mbf=\calR_{G'}(\mbf)$\;
\label{a:mainalg}
\end{algorithm}

\section{Experiments}
\label{sec:experiments}
To demonstrate our algorithm performance, we consider examples of estimating continuous values in human body shape and pose analysis: the MoCap database \cite{BaakMulBha11} of optical motion capture data and the CAESAR human body database \cite{CAESAR1999}. For comparison, we performed experiments with existing graph Laplacian (\emph{Lap})~\cite{Lux07,BelNiy03} and iterated graph Laplacian (\emph{i-Lap})~\cite{ZhoBel11} regularizers.

\paragraph{Toy example.}
We uniformly sample $10,000$ data points in $[-1,1]\times [-1,1]$. Five points (four corners and center) were assigned labels in $\{-1,10\}$ (red dots~in Fig.~\ref{f:toy2d}). While the original graph Laplacian (\emph{Lap}) produces a ``spiky'' function, the iterated graph Laplacian (\emph{i-Lap}) and our regularizer (\emph{LG}: local Gaussian) produced smooth functions, which demonstrate the importance of high-order regularization.

\begin{figure}
\centering
\includegraphics[width=\linewidth]{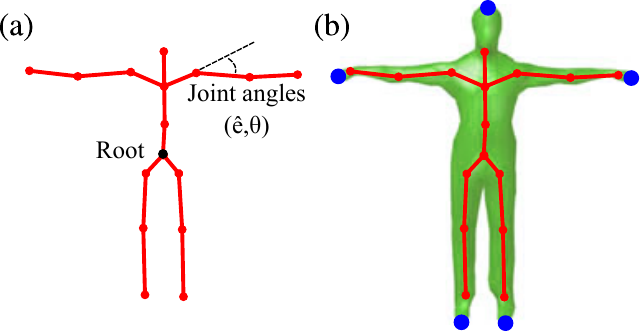}
\caption{(a) Skeletal kinematic chain. (b) End effectors (blue) recovered from a geometric model fit to the skeleton. Each joint angle is in angle-axis form, with axis $\hat{e}$ and angle $\theta$.}
\label{fig:mocap}
\end{figure}

\paragraph{MoCap database.}
This contains $50,000$ entries describing human body poses captured with an optical marker-based system \cite{BaakMulBha11}. For each \emph{pose} entry, inverse kinematics is applied to recover skeletal joint angles represented as axis-angle ($\hat{e},\theta$). A body model comprising a surface mesh consisting of $6,449$ vertices is deformed via surface skinning by embedding this skeleton of 62 joints, leading to 42 degrees of freedom parameterized by the joint angles. The locations of end effectors (left/right hand, left/right foot, and head) were separately recorded from the surface mesh model. These constitute a $15~(5\times 3)$-dimensional coarse, mid-level representation (Figure~\ref{fig:mocap}). The task is to estimate the 42-dimensional joint angles from the mid-level representation. This is useful for retrieval and indexing of motion data, \eg, for motion capture with motion priors of similar poses~\cite{BaakMulBha11}, fast MoCap data indexing in authoring tools~\cite{MulRodCla05}, or synthesis of motions from sparse sensor data with pose priors~\cite{Tautges11}. 

We randomly chose $100$ labels, with the remaining data points used as unlabeled examples. The experiment was repeated $10$ times with different sets of labeled examples and the results were averaged (Table~\ref{t:mocap}). We also show the corresponding results measured in the $186$ $(62\times 3)$-dimensional joint location space that is restored by applying forward kinematics. Both in terms of joint angle and position error, we outperform the competing methods. 

\begin{table}[t]
\centering
\caption{Mean L2-reconstruction error on the MoCap dataset.}
\begin{tabular}{l rrr}
\toprule
Algorithm&\emph{Lap}& \emph{i-Lap}& \emph{LG}\\
\midrule
Joint angles error&$1.62$&$1.24$&$\mathbf{1.16}$\\
Joint locations error&$1.22$&$0.72$&$\mathbf{0.50}$\\
\bottomrule
\end{tabular}
\label{t:mocap}
\end{table}

\paragraph{CAESAR database.}
This contains $4,258$ 3D scans of human beings, along with ground-truth measurements of their bodies obtained with calipers (Fig.~\ref{fig:caesar}). Detailed description and example usages of this dataset can be found in~\cite{CAESAR1999,HauFreBla12}. With a technique based on the work of Pishchulin et al.~\cite{pishchulin15arxiv}, we fit a statistical body model to each of the scans, which is able to represent body variations such as height, hip and belly girth, limb length, and so on. Each body scan is represented as a vector in 20-dimensional feature space spanned by a linear shape basis.

Table \ref{t:caesar} shows absolute error in semi-supervised learning performance when comparing the three regularizers, over different numbers of labeled items. Each experiment was repeated 10 times and averaged. In most cases, our approach significantly improves performance. The worse performance of \emph{LG} over \emph{i-Lap} for some cases is caused by over-fitting in cross-validation.

\begin{table*}[t]
\centering
\caption{Mean absolute error for estimating 6 ground truth parameters from the CAESAR dataset. Bold face marks the best results. The \emph{Deviation from mean} replaces the evaluation of each $f(X_i)$ with the mean of each output variable (calculated from the entire data set). This presents an idea of the difficulty of the estimation problem for each parameter.}
\vspace{0.1cm}
\setlength{\tabcolsep}{4pt}
{
\begin{tabular}{r c rrrr rr}
\toprule
\# Labels & Algorithm& \emph{Age}& \emph{Arm length}& \emph{Shoulder breadth}& \emph{Weight}& \emph{Sit height}& \emph{Foot length}\\
\toprule
&\emph{Deviation from mean}& $10.89$ & $35.98$& $36.13$ & $13.94$ & $39.50$& $15.57$\\ 
\midrule
\multirow{3}{*}{$20$}&\emph{Lap}& $\mathbf{10.89}$ & $30.23$& $32.69$ & $12.80$ & $32.58$& $13.80$\\ 
&\emph{i-Lap}& $12.46$ & $19.54$& $25.34$ & $6.30$ & $20.54$& $10.30$\\ 
&\emph{LG}& $12.55$ & $\mathbf{17.92}$& $\mathbf{20.64}$ & $\mathbf{3.17}$ & $\mathbf{19.31}$& $\mathbf{9.87}$\\
\midrule
\multirow{3}{*}{$50$}&\emph{Lap}& $10.79$ & $24.28$& $28.88$ & $10.99$ & $26.05$& $11.14$\\ 
&\emph{i-Lap}& $\mathbf{10.61}$ & $17.43$& $21.14$ & $6.62$ & $18.39$& $\mathbf{8.20}$\\ 
&\emph{LG}& $11.03$ & $\mathbf{16.30}$& $\mathbf{16.15}$ & $\mathbf{2.25}$ & $\mathbf{16.49}$& $8.34$\\
\midrule
\multirow{3}{*}{$100$}&\emph{Lap}& $10.64$ & $20.62$& $26.00$ & $9.60$ & $21.72$& $9.46$\\ 
&\emph{i-Lap}& $10.21$ & $16.97$& $19.33$ & $5.08$ & $17.65$& $\mathbf{7.99}$\\
&\emph{LG}& $\mathbf{9.85}$& $\mathbf{15.07}$& $\mathbf{15.39}$& $\mathbf{1.98}$& $\mathbf{15.59}$& $8.05$\\
\midrule
\multirow{3}{*}{$200$}&\emph{Lap}& $10.45$ & $18.23$& $23.07$ & $8.09$ & $18.99$& $8.38$\\ 
&\emph{i-Lap}& $9.99$ & $16.49$& $17.56$ & $4.11$ & $17.25$& $7.81$\\ 
&\emph{LG}& $\mathbf{9.40}$ & $\mathbf{13.96}$& $\mathbf{14.93}$ & $\mathbf{1.77}$ & $\mathbf{12.42}$& $\mathbf{7.76}$\\
\midrule
\multirow{3}{*}{$500$}&\emph{Lap}& $10.00$ & $16.44$& $19.39$ & $6.02$ & $17.31$& $7.75$\\ 
&\emph{i-Lap}& $9.52$ & $15.62$& $15.84$ & $2.93$ & $16.65$& $7.59$\\ 
&\emph{LG}& $\mathbf{8.93}$ & $\mathbf{13.42}$& $\mathbf{14.53}$ & $\mathbf{1.60}$ & $\mathbf{11.94}$& $\mathbf{7.54}$\\ 
\bottomrule
\end{tabular}
}
\label{t:caesar}
\end{table*}

\paragraph{Parameters.}
There are four hyper-parameters in our algorithm: the number ($k$) of nearest neighbors, the dimensionality ($m$) of the manifold, the regularization parameter ($\lambda$), and the local scale parameter ($\sigma$; see Eq.~\ref{e:inftycoeff}). In preliminary experiments, the performance of our algorithm varied significantly with respect to the first three parameters, while it was rather robust to $\sigma$ variations. We decide $\sigma$ adaptively for each point $X_i$, at $0.1$ times the mean distance between $X_i$ and the elements of $N_k(X_i)$ while the remaining three hyper-parameters were optimized by 5-fold cross-validation (CV) where, in each run, a subset of labeled points were left out while all unlabeled data points are kept. There are three and four hyper-parameters for \emph{Lap} and \emph{i-Lap}, respectively: $\lambda$, $k$, and the parameter $b$ for building the graph Laplacian (Eq.~\ref{e:lapgaussian}) for \emph{Lap} and the iteration parameter $p$ for \emph{i-Lap} (Eq.~\ref{e:iterlap}). These parameters were tuned in the same way as for $\emph{LG}$. Across Table \ref{t:caesar}, $k$ varied from 20 to 40, $m$ from 10 to 17, $\lambda$ from $10e^{-8}$ to $10e^{-5}$, $b$ from $5$ to $300$, and $p$ from $1$ to $4$. 

\paragraph{Computation complexity and time.}
For each algorithm, this depends on the number of data points $u$, the number of nearest neighbors $k$, and the number of non-zeros entries of the resulting regularization matrix that lies in-between $O(uk)$ and $O(uk^2)$, depending on the well-behavedness of neighborhoods (where $O(uk^2)$ corresponds to random neighbors). The most time-consuming component of each algorithm is solving the corresponding system. 

For the MoCap dataset, with $u=50,000$, $k=20$, and $p=4$ for \emph{i-Lap}, it took $30$, $50$, and $40$ seconds for \emph{Lap}, \emph{i-Lap}, and \emph{LG} to build the regularization matrices, respectively. The corresponding sparsity, defined as the number of nonzero entries divided by the number of all entries in the regularization matrix, is 0.0005, 0.0400, and 0.0017 for \emph{Lap}, \emph{i-Lap}, and \emph{LG}, respectively. This resulted in the run-times for solving the systems of roughly $50$, $720$, and $120$ seconds, respectively, on an Intel Xeon 3GHz CPU in MATLAB. For the CAESAR dataset, with $u=4,258$, run-times were only a few seconds. The improvement in computation time for large sets, coupled with the accuracy improvements demonstrated, makes our new regularizer a good alternative to \emph{Lap} and \emph{i-Lap}.

\section{Discussion}
We focused on constructing analytic solutions of Eq.~\ref{e:learningproblem}. In general, an iterative solver can be used instead (\ie, gradient descent). In this case, the iterated Laplacian \emph{i-Lap} need not be computed explicitly as its action on a vector can be computed by iterating matrix-vector multiplications. We briefly explored this possiblity: During gradient evaluation, the number of matrix-vector multiplcations increases from 1 to $p$: For MoCap ($u$=50,000, $p$=4), \emph{i-Lap} iterative optimization was around five times slower than analytic optimization, and three times slower than our iterative LG optimization. For \emph{i-Lap} with $p>4$, analytic optimization is not feasible and the iterative \emph{i-Lap} could be used; however, our LG requires no iteration. This suggests that LG can still be faster than \emph{i-Lap}. For larger-scale problems, both methods need iteration.

Local first-order approximation approaches, like ours, are supported by their success in manifold learning and regularization~\cite{RowSau00,DonGri2003}. However, local first-order approximations result in the corresponding derivatives being exact up to second order, but at third order and higher, the derivatives may deviate from the underlying covariant derivatives. Nevertheless, since the highest-order terms agree, calculating the Euclidean derivatives therein enables us to completely eliminate the possibility of generating degenerate functions. 

Furthermore, the number of hyper-parameters to be tuned (the other parameter $\sigma^k$ is adaptively decided) is the same as for classical graph Laplacian and is one smaller than for iterated graph Laplacian. Combined with the observed empirical performance of our algorithm, and the computationally efficient regularization, this supports its usage. 

Our local Gaussian interpolation varies $\sigma^k$ with the local neighborhood size $N_k(X)$ (instead of making it constant per dataset), which desires rigorous limit case behavior analysis. Further future work should address the theoretical analysis of our regularizer (\eg, error bound), and the possible benefit to spectral clustering and dimensionality reduction.

\section{Conclusion}
We have presented the local Gaussian regularizer: a new high-order regularization framework on data manifolds. Our algorithm does not suffer from the degeneracy of graph Laplacian-based regularizers. Further, it leads to a sparse regularization matrix, thereby facilitating application to large-scale datasets. Experiments on human body shape and pose analysis demonstrate the improved accuracy and faster execution time of our new algorithm.

\section*{Acknowledgements}
This work has been benefited from discussions with Matthias Hein, and from the dataset processing and model fitting work of Leonid Pishchulin and Thomas Helten. Kwang In Kim thanks EPSRC EP/M00533X/1 and EP/M006255/1, James Tompkin and Hanspeter Pfister thank NSF CGV-1110955, and James Tompkin and Christian Theobalt thank the Intel Visual Computing Institute. Part of this work was completed while Kwang In Kim and James Tompkin were at Max Planck Institute for Informatics.

\bibliographystyle{ieee}
\bibliography{./biblio}

\end{document}